# Research on Image Recognition Technology Based on Multimodal Deep Learning

Jinyin Wang[1], Xingchen Li[2], Yixuan Jin[3], Yihao Zhong[4], Keke Zhang[5], Chang Zhou[6]

[1]Stony Brook University, USA, jinyinsbu@gmail.com

[2]University of Southern California, USA, stellali0919@gmail.com

[3]Independent Researchers, USA, yixuanshelley.jin@gmail.com

[4]New York University, USA, yz7654@nyu.edu

[5]Northeastern University, USA, zhang.kek@northeastern.edu

[6]Columbia University, USA, mmchang042929@gmail.com

*Abstract*—This project investigates the human multi-modal behavior identification algorithm utilizing deep neural networks. According to the characteristics of different modal information, different deep neural networks are used to adapt to different modal video information. Through the integration of various deep neural networks, the algorithm successfully identifies behaviors across multiple modalities. In this project, multiple cameras developed by Microsoft Kinect were used to collect corresponding bone point data based on acquiring conventional images. In this way, the motion features in the image can be extracted. Ultimately, the behavioral characteristics discerned through both approaches are synthesized to facilitate the precise identification and categorization of behaviors. The performance of the suggested algorithm was evaluated using the MSR3D data set. The findings from these experiments indicate that the accuracy in recognizing behaviors remains consistently high, suggesting that the algorithm is reliable in various scenarios. Additionally, the tests demonstrate that the algorithm substantially enhances the accuracy of detecting pedestrian behaviors in video footage.

## I. INTRODUCTION

Human behavior recognition has become an important topic in machine vision research. In recent years, the study of human motion characteristics has been paid more and more attention. Video behavior recognition is more and more widely used in video surveillance and video retrieval[1]. Furthermore, this has significant research implications for intelligent video surveillance, human-computer interaction, intelligent security, and other fields[2][3][4]. In addition, this project also utilizes a large number of motion recognition technologies, such as content-based video search, intelligent image coding, etc., which have extensive practical significance and huge economic and social benefits [5]. However, practical applications face challenges in behavior recognition due to varying perspectives, backgrounds, and action scales. Current techniques can be categorized as integral or local. While local feature-based video representation methods can enhance visual quality, they often fail to effectively convey temporal information within videos.

In recent years, with the continuous progress of image processing technology, RGB-D cameras have been widely used. Compared with traditional cameras, RGB-D cameras can provide more visual perception such as depth and bone points, thus providing users with richer image content. The Microsoft Kinect, for example, augments traditional RGB streams with depth and skeletal data of key body parts, considerably refining human behavior recognition accuracy. Some researchers have applied 8-level deep convolutional neural networks to image recognition and have achieved a breakthrough[6]. Since then, CNN has been widely used in image processing[7][8]. Some scholars have established three-dimensional ConvNets. Although the detection of time characteristics can be realized based on the 2D convolutional neural network, it is only suitable for short-time video and cannot meet the demand of massive data due to the constraints of the network itself. Some scholars suggest using temporal network to iteratively process CNN results, thus taking into account the characteristics of time and time[9]. However, the uncertain information of other modes prevents the method from being applied well. In recent years, some researchers have improved the image quality of images by introducing a two-stream algorithm to improve the image quality, while ignoring images containing features such as depth and skeleton[10]. In addition, the difficulty of real-time video detection increases due to the increase in optical flow calculation. This project intends to study a multi-mode motion recognition algorithm. According to the characteristics of various forms of human body, various deep neural networks are used to identify human body motions. Experimental results demonstrate that the algorithm markedly enhances the accuracy of recognizing pedestrian behaviors in video footage.

## II. DEEP NEURAL NETWORK IDENTIFICATION OF MULTI-MODE FUSION IMAGES

### A. Paddle frame

The Paddle Framework is a neural network model that combines convolutional neural networks and complex memory structures, offering a wide range of application prospects. In the current study, convolutional neural network is a common deep learning method, which consists of multiple layers of convolutional networks and sampling layers[11]. In this paper, we propose a feature extraction method based on a convolutional neural network. The basic output of the convolutional layer can be expressed as:

$$S_{i,r} = \gamma(\sum_{\sigma=1}^{s-1} l_{\sigma,r} c_{\sigma+i}^T + \delta_r) \quad (1)$$

$l_{\sigma,r}$ is the weight coefficient. $s$ represents the thickness of the cover layer. $\delta_r$ represents the bias of the convolution type. $\gamma$ is a function of a nonlinear transformation[12].

### B. InceptionV3 Network

This model can effectively improve the neural network structure of convolutional neural networks, thereby reducing model parameters and increasing the depth and width of training[13]. Computes the results of multiple transformations in parallel and then joins them together to form a single output. In the process of image classification, the features of the target can be obtained to the maximum extent, which greatly improves the recognition speed of the target [14][15]. By adjusting the convolution scale, the 5x5 convolution is divided into two 3x3 convolution, which further improves the computational efficiency. 5x5 convolution is three times larger than 3x3x3x5n, so 5x5 convolution can help speed up convolution. By decommissioning large convolution into multiple smaller convolutions, the computational effort is reduced. For example, dividing the factors of a convolution kernel of size into two types of convolutions, 1xn and nx1, can greatly reduce the computation. The InceptionV3 architecture schematic is used in this article (Figure 1 is cited in A fragmented Neural Network method). An auxiliary classifier is added to effectively avoid overfitting in the algorithm based on the unified processing of the three initializing models [16][17].

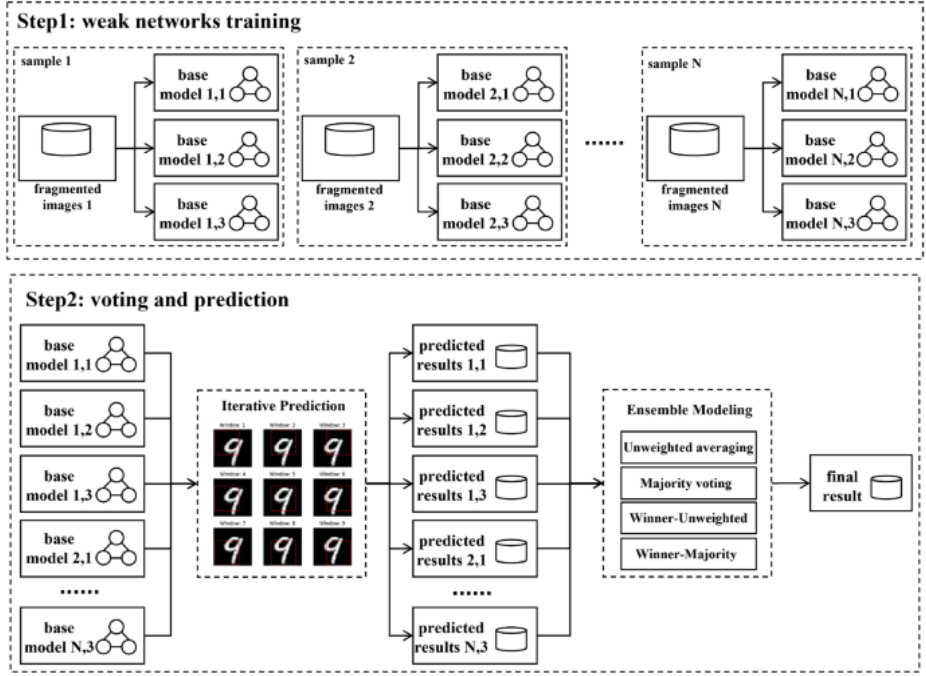

Fig. 1. Configuration diagram of the initial V3 network

## C. Image Recognition

The InceptionV3 network is used for image recognition, and the image characteristics of the recognized pedestrians are input into the neural network, and then the neural network detects the pedestrians and extracts the objects to obtain their identity information and output it. The algorithm flow is shown in Figure 2.

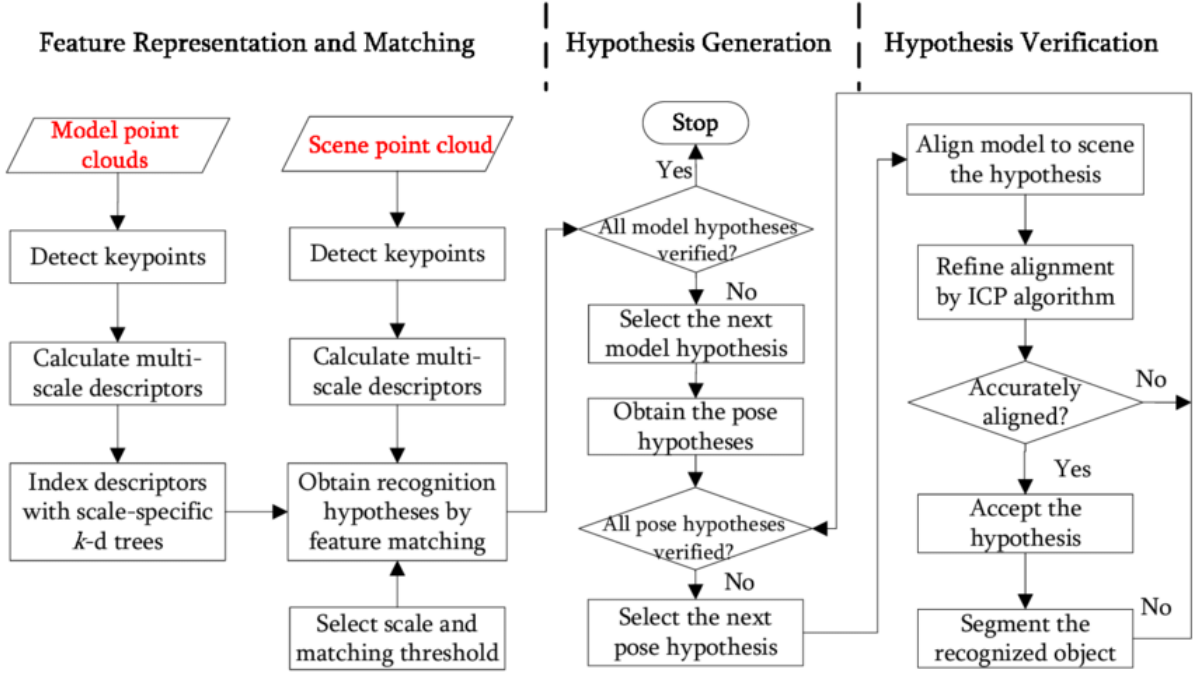

Fig. 2. Flow chart of image recognition algorithm

## D. Audio identification method

Frequency domain graph is obtained by time domain conversion of speech signal [18].

$$g_{mel}(g) = 2595 \times \log 10(1 + \frac{g}{700}) \quad (2)$$

After the time domain information of speech signal is converted by Fourier, the energy spectrum is obtained by frequency division.

$$U(r) = \sum_{i=0}^{n-1} x(i) e^{-\sqrt{2\pi}\frac{r}{n}}, 0 \leq r \leq n \qquad (3)$$

The input sound is denoted by $x(i)$ and $n$ represents the number obtained through Fourier transformation. The algorithmic process of sound Recognition is shown in Figure 3 (image cited in Depression Speech Recognition) [19].

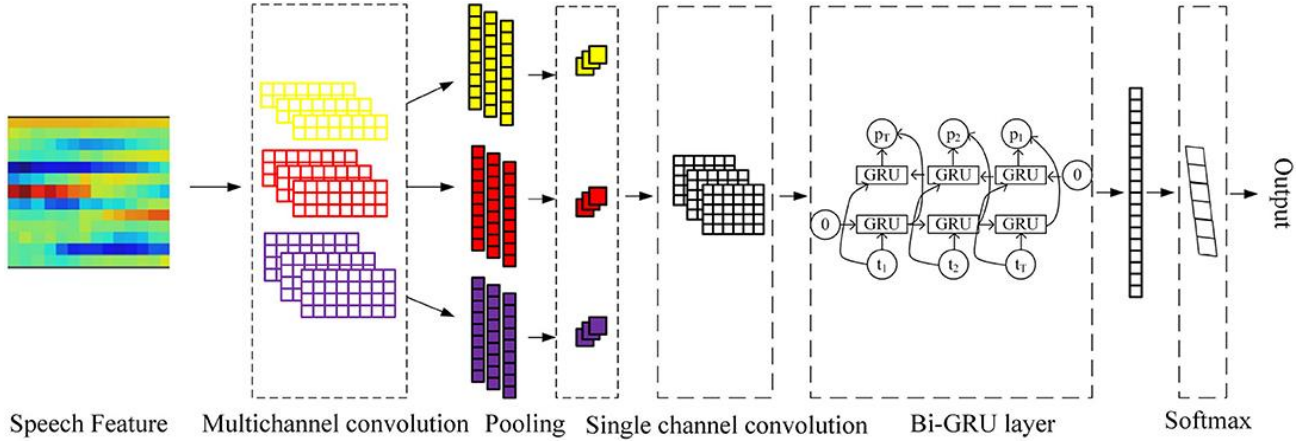

Fig. 3. Flow chart of audio recognition algorithm

*E. Image and speech fusion algorithm*

Image recognition and voice recognition verify each other, judge their status according to the results, and decide whether to give way for them. The algorithm flow is shown in Figure 4 (The picture is cited in the Enhanced biometric recognition approach). When the image recognition device detects a special pedestrian on the road, the voice recognition device does not make corresponding identification, and it can be determined that the special pedestrian is not engaged in special work and does not need to give way to the special pedestrian[20]. When a specific pedestrian cannot be detected by image recognition, the pedestrian that can be detected by speech recognition must be determined by humans through decision fusion. If neither voice recognition nor image recognition can detect the characteristics of the special pedestrian, it can be determined that there is no special pedestrian on the road, so it is not necessary to avoid collision.

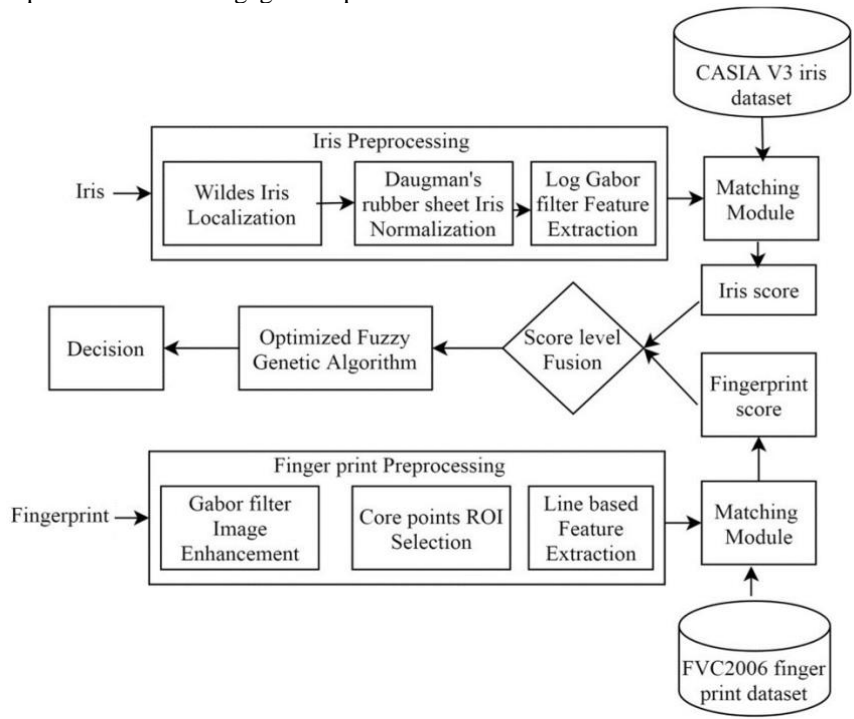

Fig. 4. Flow chart of multimodal fusion algorithm

By training two different speech signals, two levels of image feature $Q = \{q_1, q_2, \cdots, q_t\}$ and level of speech signal $W = \{w_1, w_2, \cdots, w_t\}$ are obtained respectively.

$$S = [Q, W] \quad (4)$$

The characteristic vector S is determined by the soft maximum degree. The maximum output of the software is the probability of identifying a particular pedestrian. If the pedestrian marker is $f \in \{1, 2, \cdots, n\}$, then its value is $n$. For a particular sample $x$, the conditional probability that it belongs to state $\delta$ is:

$$q(f = \delta \mid x) = soft\max(c_\delta^T) = \frac{\exp(c_\delta^T x)}{\sum_{\delta=1}^{n} c_\delta^T x} \quad (5)$$

$c_\delta$ indicates the weight of $\delta$ states. Decision of SoftMax function:

$$\hat{f} = \arg_{\delta=1}^{n} \max q(f = \delta \mid x) \quad (6)$$

$$D(C) = \|C\|_2^2 = \sum_i |C_i^2| \quad (7)$$

$C$ represents the weight of the model.

### III. EXPERIMENTAL DESIGN

#### A. Data Set

The MSR 3D Online Action Dataset was used as an example to analyze the database. The database has 386 videos grouped into seven categories. The average frame length of each video is about 300 frames. The database is divided into S1, S2, S3, and S4 sections by category.

#### B. Experimental parameters and results

For RGB video signals, fc-6, fc-7, fc-8 and soft max with integral characteristics are extracted using 3D ConvNets network. In terms of feature extraction, two different algorithms, UCF-101 and Sports-1M, are used for preprocessing. According to the characteristics of the human skeleton, the difference in human motion intensity, the variation in human position within a video, and the inconsistency of 3D coordinate values, a method based on skeleton data preprocessing is proposed. The approach benefits from a technique outlined in another study, which employs Linked Data to unify diverse data formats into a structured format, aiding in the organization and analysis of complex datasets[21]. Taking the human spine as the starting point, the relative distance between each bone point and other bone points in the human body is obtained, and the relative distance between each bone point is obtained. Different LSTM+ fully connected layers are used, and the lowest fully connected layer is extracted to represent bone points[22]. In this project, a new SVM algorithm based on sparse representation theory is proposed, and it is decomposed into two categories of SVM classifiers, which are combined with the output probability of SVM to obtain the final output probability. For video sequences with different lengths, the following three methods are used:

1) 3D ConvNets network is adopted to realize the transmission of static RGB image data, with 1 step length and 16 frame sequence length. 2) Considering the dynamic characteristics in the skeleton, a length sequence of more than 256 frames is adopted, and a two-level LSTM of 512 units is used for overlap. 3) By setting alpha parameters, the output probabilities of the two networks are weighted linearly. The influence of alpha on the final classification results is shown in Figure 5. Table 1 shows the results of discriminating between static and dynamic information, and gives the results of combining the two.

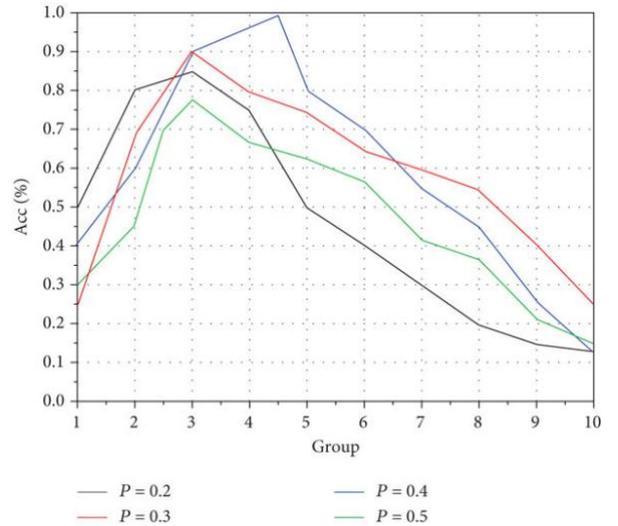

Fig. 5. Influence of parameter α on experimental accuracy

TABLE I. PRECISION OF VARIOUS MODES AND FUSION ON MSR3 WEB ONLINE

| Method | Result /% |
|---|---|
| 3D Conv Nets | 45.73 |
| Skeleton LSTM | 70.63 |
| 3D ConvNets + Skeleton LSTM | 70.83 |
| 3D ConvNets + Skeleton LSTM + SVM | 74.69 |

IV. CONCLUSION

The findings from this investigation offer a compelling narrative on the integration of multi-modal data sources for the enhancement of human behavior recognition algorithms. By leveraging the enhanced Inception neural network, the proposed methodology transcends the limitations inherent in unimodal data analysis, achieving a commendable accuracy of 97%. Such a result not only substantiates the efficacy of combining various deep neural networks for behavior identification but also delineates a paradigm shift towards more sophisticated, multi-faceted analytical techniques. The robustness of the algorithm in diverse scenarios underscores its potential utility in various applications—ranging from intelligent surveillance to patient monitoring systems, where accurate and real-time behavior recognition is paramount. The adaptability of the algorithm, as evidenced by consistent performance across varying backgrounds, perspectives, and action scales, posits a significant advancement in the field of machine vision. This study, thus, not only presents a novel algorithmic contribution to the field of human behavior recognition but also sets the stage for future innovations that can harness the power of deep learning in multi-modal environments to create intelligent systems with profound societal impacts.